\documentclass{article}
\usepackage[letterpaper, total={6in, 8in}]{geometry}
\usepackage{comment}
\usepackage{amsmath,amssymb,amsfonts}
\usepackage{algorithmic}
\usepackage{graphicx}
\usepackage{textcomp}
\usepackage{xcolor}
\usepackage{subcaption}

\DeclareMathOperator*{\argmax}{arg\,max}

\usepackage{multirow}

\newcommand\Tstrut{\rule{0pt}{2.2ex}} 

\title{\LARGE \bf
Interpretable Identification of Comorbidities Associated with Recurrent ED and Inpatient Visits 
}
\date{}

\author{Luoluo Liu$^{1}$, Dennis Swearingen$^{2}$, Eran Simhon$^{1}$, Chaitanya Kulkarni$^{3}$, David Noren$^{1}$, Ronny Mans$^{4}$
\thanks{$^{1}$Luoluo Liu is with Philips Research North America (PRNA), Cambridge, USA. Email: {\tt\small luoluo.liu@philips.com}}%
\thanks{$^{2}$D. Swearingen, MD is with Banner Health, Phoenix, USA. $^{1}$ E. Simhon, D. Noren are with PRNA; $^{3}$ C. Kulkarni is with Philips Innovation India; $^{4}$ R. Mans is with Philips Research Europe, Netherlands.  
}}

\begin{document}

\maketitle
\thispagestyle{empty}
\pagestyle{empty}

\begin{abstract}

In the hospital setting, a small percentage of recurrent frequent patients contribute to a disproportional amount of healthcare resource utilization. Moreover, in many of these cases, patient outcomes can be greatly improved by reducing re-occurring visits, especially when they are associated with substance abuse, mental health, and medical factors that could be improved by social-behavioral interventions, outpatient or preventative care. Additionally, health care costs can be reduced significantly with fewer preventable recurrent visits.

To address this, we developed a novel, interpretable framework that both identifies recurrent patients with high utilization and determines which comorbidities contribute most to their recurrent visits. 
Specifically, we present a novel algorithm, called the minimum similarity association rules (MSAR), which balances the confidence-support trade-off, to determine the conditions most associated with re-occurring Emergency department and inpatient visits.  We validate MSAR on a large Electronic Health Record dataset, demonstrating the effectiveness and consistency in ability to find low-support comorbidities with high likelihood of being associated with recurrent visits, which is challenging for other algorithms such as XGBoost.
\newline
\indent \textit{Clinical relevance}— In the era of value-based care and population health management, the proposal 
could be used for decision making to help reduce future recurrent admissions, improve patient outcomes and reduce the cost of healthcare.

\indent \textit{Index Terms}--- Recurrent patients, Emergency Department, inpatient  re-admissions, confidence-support trade-off, Association Rules
\end{abstract}


\section{Introduction}
Recurrent patients, also known as "frequent flyers", "high utilizers", or "super users" in hospitals, are a small group; however, they impose a disproportionately high utilization of resources 
\cite{onepercent} \cite{ng2019characterization}. In fact, the top one percent contribute to $22\%$ of health care spending, and the top $5\%$ account for around $50\%$ of overall costs \cite{onepercent}. Further, the top $15\%$ of utilizers contribute to around $85\%$ of total health care costs \cite{ng2019characterization}.
Accordingly, identifying potentially preventable visits among this group of patients could reduce hospital and health care costs significantly.

There are two major types of recurrent patients. One type includes those with mental health or substance (drug and/or alcohol) abuse conditions.  It has been shown that these patients contribute to high utilization of Emergency Departments (EDs) \cite{smith2015hospital,doupe2012frequent,crane2012reducing,owens2010mental,pines2011frequent,larkin2005trends} and inpatient visits in community hospitals \cite{smith2015hospital,stranges2011state}. 
In 2008, mental health and substance abuse (MHSA) were the principal reasons behind $1.8$ million patient hospitalizations, accounting for $4.5\%$ of all hospitalizations in the U.S and costing $\$9.7$ billion
\cite{stranges2011state}. Additionally,  a large percentage of MHSA  patients lack insurance coverage, are sometimes unemployed, and have low average education levels \cite{o2006managing}\cite{kangovi2014patient}.  Fortunately, it has been reported that 
rehabilitation programs \cite{mckay1994treatment}, accommodations \cite{o2006managing}, social support in the community \cite{martinez2019time} and other outpatient programs \cite{o2006managing} improve patient outcomes and reduce 
re-admissions.  

A second type of potentially preventable recurrent admission relates to chronic diseases, such as Acquired Immunodeficiency Syndrome (AIDS) and diabetes mellitus. This type of recurrent visits could be reduced through proactive planning and care. 
About half of AIDS 
re-admissions could be prevented by providing better access to medical supplies, assuring adherence to follow-up tests and visits, improving treatment compliance, and satisfying patients' psychological and social needs \cite{nijhawan2015half}. 
Similarly, 
robust diabetes management, discharge planning, and post-hospital instructions could lower ED visits for diabetes patients \cite{ostling2017relationship}.

One of the biggest challenges in addressing recurrent visits is the lack of a standard criteria for identifying these patients. In fact, 
over $180$ criteria from about $100$ sources are surveyed 
\cite{grafe2020classify}. 
An existing work combines $180$ previously proposed rules using a clustering algorithm \cite{grafe2020classify}. As a result, the final set of rules takes many factors, across clinical and operational elements, into account. This procedure 
makes the algorithm very complex and lacking in interpretability, creating a barrier for practical usage. 

Due to the lack of standard criteria for identifying recurrent patients,
our first aim is to identify this group of patients. 
Our second aim is to develop an interpretable method to provide insight into the pertinent factors contributing frequent visits. In the Electronic Health Record, the underlying reason for a particular patient visit can be inferred from the chief complaint or directly obtained if documented. 
However, for recurrent patients, 
the 
reason for any given visit might not be representative enough to explain the recurrent visits.

In identifying MHSA and chronic conditions associated with high utilization, the third key challenge is that some conditions (for example drug abuse, AIDS) have \textbf{low prevalence/support} but potentially \textbf{high likelihood/confidence} compared to others (such as hypertension with low confidence but high support). It is difficult for conventional machine learning methods such to detect those conditions, which is validated in Section \ref{subsec:xgboost}. Aiming at developing a explainable model, we investigated on 
a interpretable statistical method called Association Rules \cite{agrawal1993mining} (AR). The conventional AR method takes the maximum confidence/likelihood from all candidate rules exceeding a minimum support/prevalence \cite{agrawal1993mining}. Even though AR is advantageous due to its high interpretability, 
the main drawback is:
\textbf{a conventional AR method with a fixed minimum support threshold could potentially eliminate those high confidence but low support conditions}.

Variants of AR methods have been developed for balancing confidence and support to help resolve this issue. 
Authors in \cite{rudin2013learning} propose a modification on confidence calculation by adding an extra parameter. 
However, requiring the additional term makes it hard to interpret in deployment. 
Authors in \cite{scheffer2001finding}
use prediction of future data to determine optimal weights. The underlining assumption that the distribution of comorbidities associated with frequent visits are consistent over time, is
not likely 
true especially if some preventative efforts has been taken over the years.

Instead, we propose an algorithm named minimum similarity association rules (MSAR) that improves on the conventional AR method by \textbf{balancing the support and confidence trade-off} with optimal weights learned from data which are easily adjustable to varying deployment sites.
In addition, the MSAR algorithm for identifying 
top  comorbidities associated with recurrent visits is \textbf{highly interpretable, efficient, consistent, customizable, and deployment-friendly}.  
%
The proposed algorithm is particularly well suited to handle cases with a large variance in support distribution, which is crucial given that comorbidities vary widely in prevalence.  Additionally, MSAR successfully identifies challenging case of \textbf{low-support comorbidities with high likelihood of recurrent visits}. The comparison between different algorithms is listed in Table~\ref{table:summary}.

\begin{table}[h!]
  \begin{center}

    \begin{tabular}{|l|c|c|c|}
    \hline
      \multirow{2}{*}{Algorithms} & Association & XGBoost\cite{chen2016xgboost} & MSAR\\
 &Rules (AR)& +Shapley\cite{shapley} & {(Ours)}\\
      \hline
      \hline
      interpretability
 & High & Medium & \textbf{High}\\
 \hline
\Tstrut
 Consistency&High &Medium & \textbf{High}\\
 \hline
      balances conf-supp & \multirow{2}{*}{No} & \multirow{2}{*}{No} & \multirow{2}{*}{\textbf{Yes}}\\ trade-offs &  &&\\
      \hline
      ability to select&\multirow{2}{*}{Limited}& \multirow{2}{*}{Limited} & \multirow{2}{*}{\textbf{Yes}}\\ high-conf, low-supp factors &  &  & \\
 \hline

ability to distinguish&&&\\ across factors 
 &Limited & Limited & \textbf{Yes} \\
 \hline
    \end{tabular}
  \end{center}
      \caption{ Algorithm comparisons for selecting top factors.}
    \label{table:summary}
\end{table}

\section{Method}
\begin{figure*}
    \centering
    \includegraphics[width = 16cm, height = 3.5cm]
    {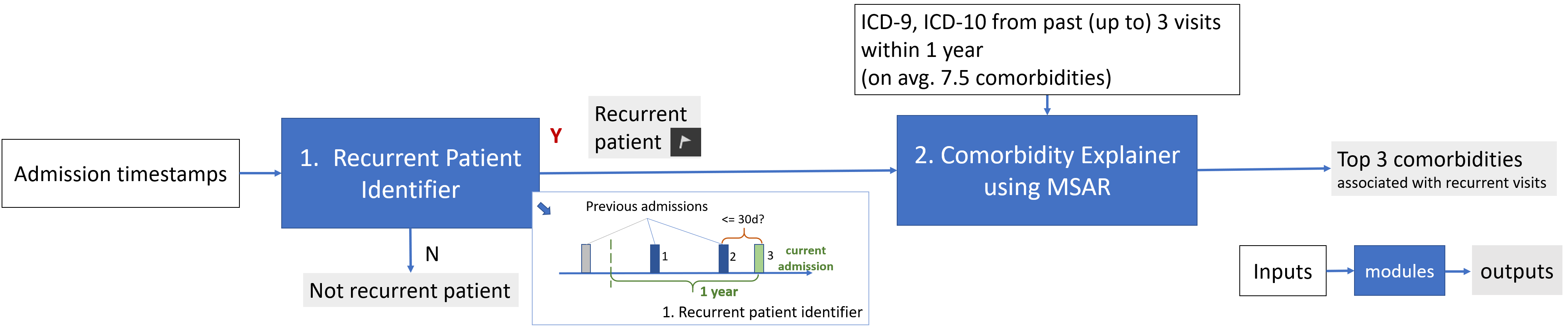}
    \caption{The first module identifies a recurrent patient and the Comorbidities Explainer outputs the top $3$ comorbidities associated with frequent visits for that patient. Rules are trained on retrospective data with both recurrent and non-recurrent patients. The algorithm can be executed for recurrent patients on an individual patient level.}
    \label{fig:framework}
\end{figure*}

The flowchart of the proposal is illustrated in Figure~\ref{fig:framework} with two corresponding modules. The recurrent patient identifier 
module uses 
admission timestamps within $1$ year to determine the frequency of admission, and then determines whether the visitor is a recurrent patient. 
The  comorbidities explainer module, gives the top three comorbidities associated with their frequent visits.


\subsection{Recurrent Patients Identifier}

One of the challenges in reducing preventable recurrent visits is the lack of a standard definition for recurrent patients. 
Previous works have defined recurrence based on utilization, physiological states, and cost 
factors~\cite{grafe2020classify}. Among these choices, utilization elements are the most easily accessible from the EHR, and the algorithm can be executed at the time of admission. We therefore decided to use top utilization data elements from the literature and then combine them. 

Our proposed framework identifies recurrent patients by satisfying at least one of the following criteria:
$i)$ readmission within $30$-days, which is accepted by the Centers for Medicare and Medicaid Services\cite{scmm} in measuring operational performance;
$ii)$ more than $4$ non-elective inpatient visits within a year ($44\%$ out of $180$ criteria reviewed in \cite{grafe2020classify}); 
$iii)$ more than $4$ emergency department visits within $1$ year  ($41\%$ out of $180$ criteria reviewed in \cite{grafe2020classify}). 

\subsection{Comorbidity Explainer via Min-similarity Association Rules (MSAR)}

We introduce the proposed algorithm: Min-similarity Association Rules (MSAR), 
for selecting the top comorbidities associated with recurrent visits. This algorithm seeks optimal solutions that balance confidence and support, learned from retrospective data.  It also addresses the difficulty in identifying high-confidence but low-support comorbidity combinations.
Below we describe key parts related to MSAR.

\subsubsection{Derivation of 
Comorbidities}

From the Electronic Health Records (EHR) database, we first take International Classification of Diseases (ICD)-$9$ and ICD-$10$ diagnosis codes
, which, in total, 
contain approximately $83000$ codes \cite{wikiicd}. 
To reduce the dimensionality when encoding such a large number of diagnosis codes, we use the Elixhauser comorbidity index \cite{elixhauser1998comorbidity} to map ICD codes into comorbidity categories. 
Elixhauser comorbidity is a good choice 
because it is an updated version of the widely adopted Charlson comorbidity index \cite{charlson1987new} \cite{deyo1992adapting} with newly added categories for mental disorders, drug and alcohol abuse, obesity, and weight loss, which are the factors that could potentially be improved via social-behavioral interventions and outpatient care. 
Since most recurrent patients has more than $4$ visits within a year, we take ICD codes from previous $3$ visits within $1$ year and map them to Elixhauser comorbidities using the Hcuppy $0.0.7$ \cite{hcuppy} Python package. 

\subsubsection{Number of comorbidities for recurrent patients} 
On average, recurrent patients have 
$7.5(\pm 3.7)$
comorbidities from their past three visits within a year. Displaying all comorbidities associated with recurrent visits is not efficient for visualization nor for identifying the root cause for recurrent admissions. Therefore, this motivates us to develop a comorbidity explainer to pick the top comorbidities associated with frequent visits.

\subsubsection{Confidence and support}
In the context of our problem,
\textbf{confidence} is the \textbf{likelihood of being a recurrent patient given a specific combination of comorbiditities}
; and \textbf{support} is the \textbf{prevalence of a given comorbiditity combination}. 
We first visualize the confidence and support of a single comorbidity for all types of comorbidities in Figure \ref{fig:scatter_plot_prev_supp}.
Drug abuse has the highest confidence overall. Additionally, the variance of support across different comorbidities is large, with values ranging from $0.0045$ to $0.72$. 
There are comorbidities with high confidence but low support, 
such as drug abuse, psychoses, weight loss and AIDS, which posts challenges for algorithms in deciding on key factors associated with recurrent visits.  

\subsubsection{Confidence increases and support decreases with increasing number of comorbidities}
\begin{figure}
    \centering
    \includegraphics
    [width =8.5cm, height = 7.0 cm, trim={0.0cm 0 0cm 0.0cm},clip]
   {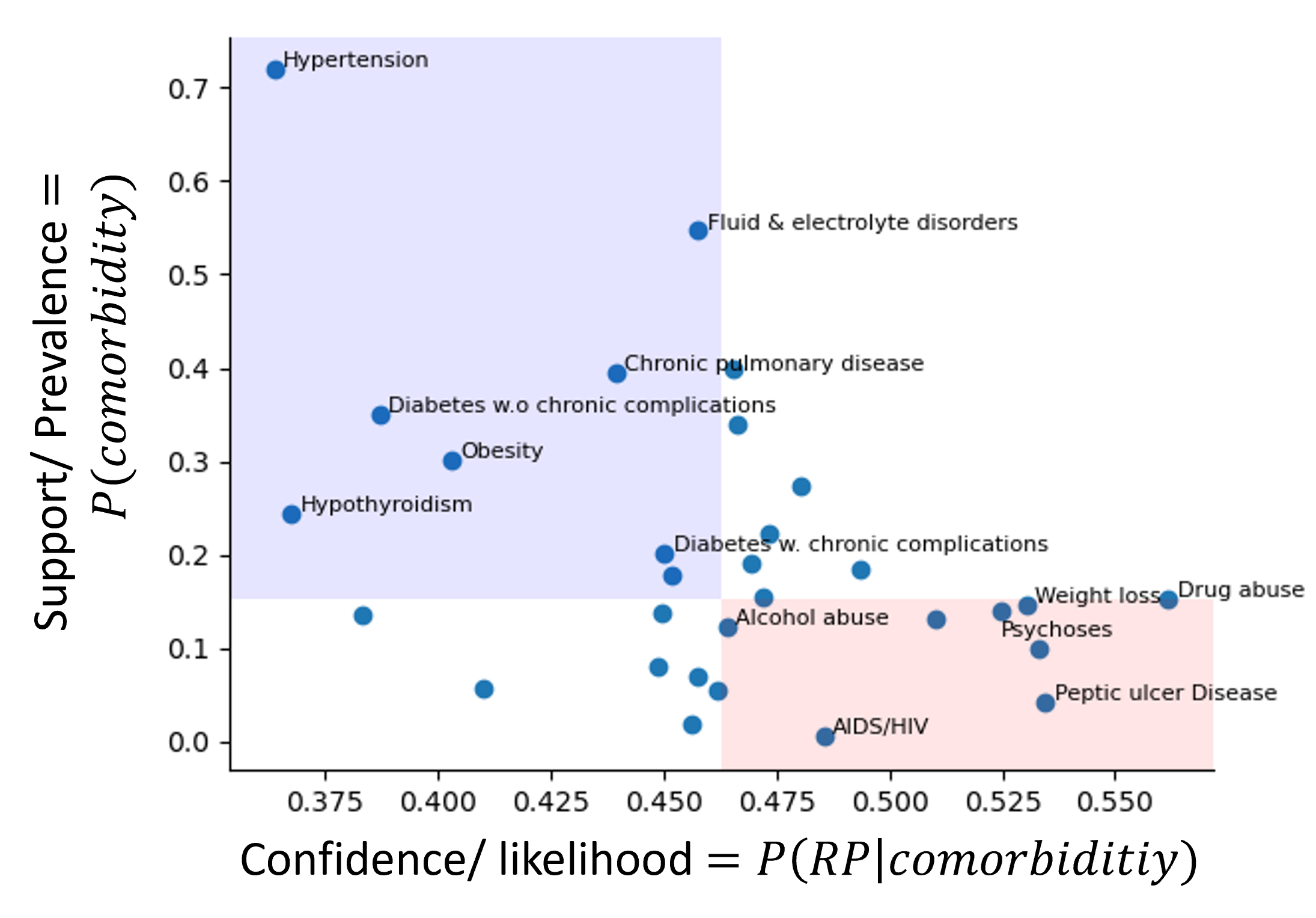}
    \caption{Confidence and support of Elixhauser comorbidities. The blue and red boundaries are medians of confidence and support, respectively. Comorbidities in the blue shaded area have low-confidence and high-support. Comorbidities in the red shaded area have high-confidence and low-support. Cohort only contains patients with 3 or above comorbidities within 1 year.}
    \label{fig:scatter_plot_prev_supp}
\end{figure}

\begin{figure}
    \centering
    \includegraphics[width = 8.8cm, height = 7.0 cm]{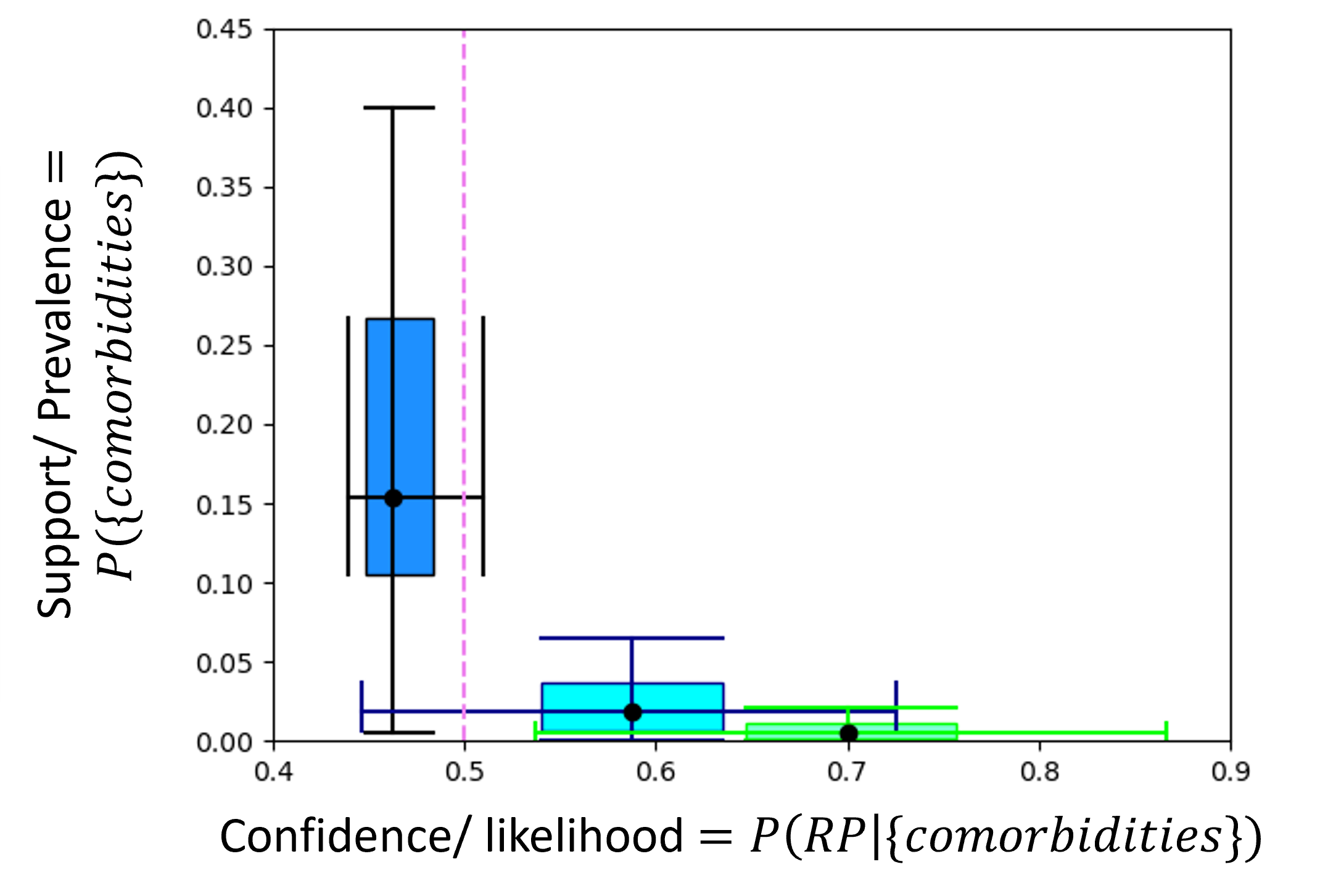}
    \caption{Increasing the number of comorbidities $n$ will increase confidence while decreasing support. Box plots show differences in ranges of confidence and support with varying numbers of comorbidities (Left $n = 1$, middle $n=2 $, right $n = 3$). Violet vertical line is a confidence marker at $0.5$.}
    \label{fig:size}
\end{figure}

From Figure \ref{fig:scatter_plot_prev_supp}, the range of confidence for  
single comorbidities is between $0.36$ to $0.56$; this range is not very wide for discriminating recurrent patients versus non-recurrent patients. To increase the discriminability,  we use a combination of comorbidities rather than a single one from past visits.  
We plot the confidence and support ranges in Figure~\ref{fig:size} to find a reasonable parameter for the number of comorbidities $n$. 
As $n$ increases, the confidence range increases, and  
 the range of supports reduces. 
 We take the comorbidities size to be $n = 3$, as the majority of their confidence are over $0.5$. 
We therefore use association rules on combinations of $3$ past comorbidities from the previous 
$\leq 3$ visits within last one year.




\subsubsection{Conventional association rules}
Our rule candidates are comorbidity combinations with $3$ different comorbidities, 
taken from ICD 
codes from the most recent 
$3$ visits. 
Thus, for a comorbidity combination of $3$ comorbidities $\{A,B, C\}$, 
the confidence$/$likelihood of recurrent patients (RP) is:
\begin{equation}
\label{eq:conf}
  \mbox{Confidence: } \  c (\{A, B, C\} ) = P(RP|A, B, C); 
\end{equation}
and its 
support$/$prevalence is: 
\begin{equation}
\label{eq:supp}
\mbox{Support: } \ s (\{A, B, C\} ) = P(A, B, C).
\end{equation}

\begin{figure}
    \centering
    \includegraphics[width = 8 cm, height = 6.0cm]{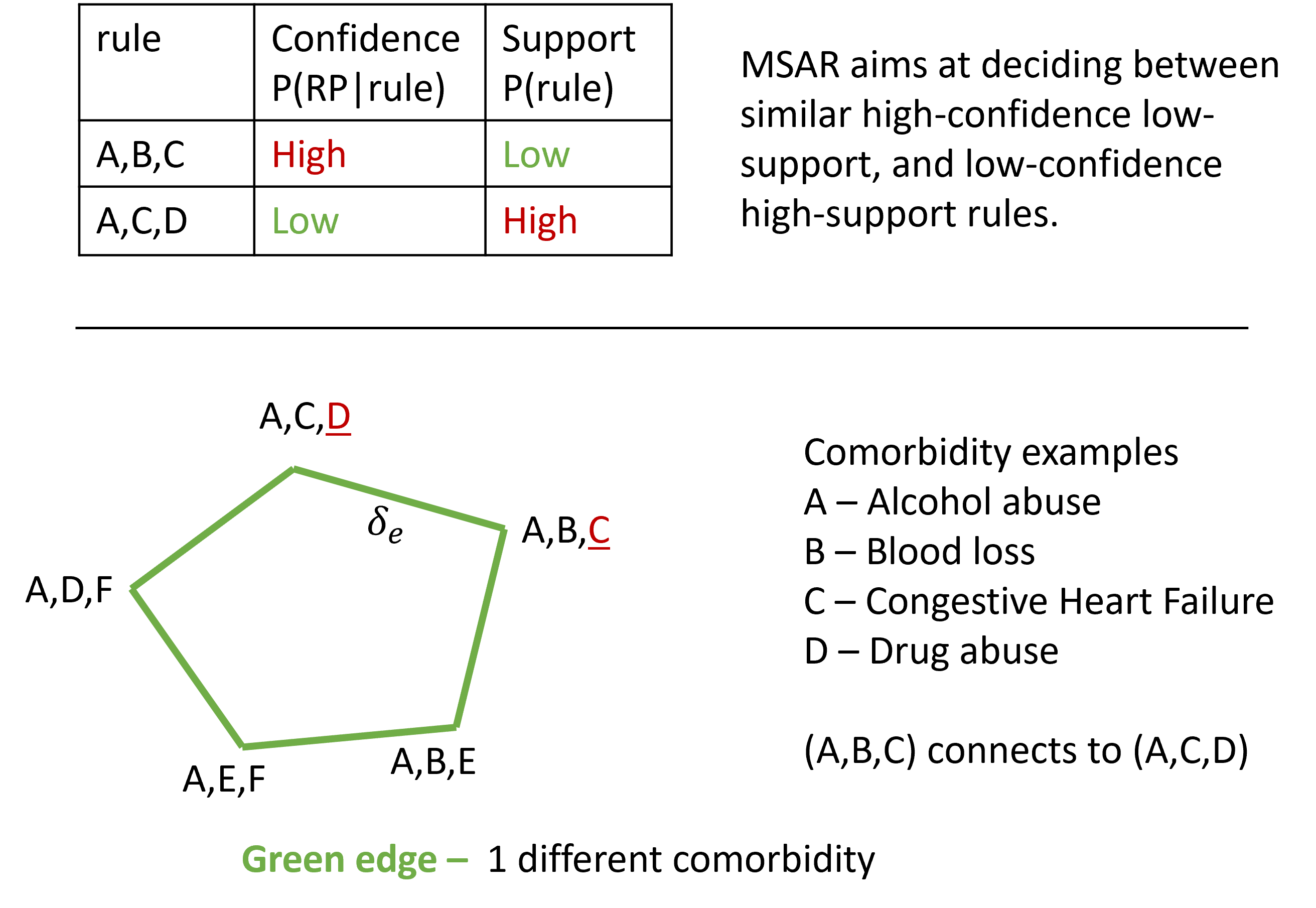}
    \caption{ 
    \textbf{Top}: The goal of MSAR is to choose between similar rules when one is higher confidence, lower support and the other is lower confidence but higher support.\\ \textbf{Bottom}:
    Illustration of min-similarity rule graph construction. The graph is built with each node being a triplet of comorbidities. An edge exists for similar rules with node pairs that differ by exactly one comorbidity. The optimization to balance confidence and support is only on pairs of similar rules with $1$ comorbidity difference. 
    }
    \label{fig:sim_rule}
\end{figure}

The conventional association rules method applies a user-specified threshold ($\tau$)  on support and then takes the highest confidence rule remaining. To simplify the notation, we use $v$ to denote a combination of comorbidities.
\begin{equation}
v^\star_{AR} = \argmax{ \mbox{Confidence} (v), \mbox{s.t. } \forall v,   \mbox{Support} (v) \geq \tau}.
\end{equation}

\subsubsection{Drawbacks of conventional AR with a fixed threshold}

Figure~\ref{fig:scatter_plot_prev_supp} and Figure~\ref{fig:size} show that there is a trade-off between confidence and support. First, in clinical applications, some diseases are 
less prevalent than others. However these may have higher confidences associated with recurrent visits, shown in Figure~\ref{fig:scatter_plot_prev_supp}. Secondly, 
while increasing the number of elements (comorbidities) can boost confidence, it largely decreases the support. 

The main drawbacks for conventional AR at a fixed threshold are:
 $i)$ A large threshold could remove rare diseases, including the ones with high confidence. $ii)$ A 
 small threshold will make the algorithm very sensitive to outliers and potentially risks revealing the identity of a patient with rare conditions.
 These drawbacks motivate us to develop an algorithm to balance confidence and support automatically.

\subsubsection{Motivation and definition of similar rules}



For similar comorbidity combinations, selecting between high-confidence or high-support rules motivates us to balance confidence-support trade-off. For example, in Table \ref{tab:msar_example}, we have two sets of comorbidities: the first is weight loss (WL), peripheral vascular diseases (PVD), hypertension, and the second set is WL, PVD, and fluid-electrolyte disorders. Those two sets have similar confidences, but they have very different supports. For a recurrent patient with all four of these comorbidities during their past three visits, the ability to differentiate which three comorbidities to prioritize as factors of recurrent visits is important, and serves as a key motivation of this solution. We therefore define pairs of similar rules and optimize the weights of confidence and support among them.



Similar rules are defined as the following: for rule set $v_1$ and rule set $v_2$, both of size $3$, they are similar rules if and only if exactly one element is different between two sets. For example, for $v_1 = \{A, B, C\}, v_2 = \{A, B, E\}, v_3 = \{A, E,F \}$, $v_1$ and $v_2$ are similar rules, whereas $v_1$ and $v_3$ are not similar rules. In graph representation, we model each rule as a vertex in a graph, 
and the edge is only present 
if there is only one 
different element between the two rule sets.
Figure \ref{fig:sim_rule} illustrates this graph structure.


\subsubsection{MSAR balances confidence and support for similar rules}

Then, we formulate the combined confidence and support rule: 
for weights $w_c$, $w_s \geq 0$, where the sum of weights is one: $w_c + w_s = 1$,  the rule score function on a vertex $v_i$ is as follows:
\begin{equation}
   r(v_i) =  w_c \Tilde{c}(v_i) + w_s \Tilde{s} (v_i),
\end{equation}
where $\Tilde{c}(v_i) , \Tilde{s} (v_i)$ are 
\textbf{$\mathbf{z-}$normalized confidence and support}, respectively.
Now, for an edge $e$ between vertices $v_i$ and $v_j$, 
we define the difference by the following: 
\begin{equation}
\begin{split}
     \delta(e) & =   r(v_i) - r(v_{j})\\
& =   w_c \Tilde{c}(v_i) + w_s \Tilde{s} (v_i) -  w_c \Tilde{c}(v_j) - w_s \Tilde{s} (v_j) \\
& =  w_c  (\Tilde{c}(v_i) - w_c \Tilde{s} (v_j) )  + w_s (\Tilde{s}(v_i) - \Tilde{s} (v_j)). 
\end{split}
\label{eq:edge_delta_ori}
\end{equation}
We use $\delta_{c} (e), \delta_{s} (e)$ to denote the difference of confidence and support between two vertices, respectively. Then, equation (\ref{eq:edge_delta_ori}) reduces to 
\begin{equation}
\begin{split}
     \delta(e)  =  w_c \delta_c(e) + w_s \delta_s(e).
\end{split}
\label{eq:edge_delta}
\end{equation}




We require similarity to be inversely correlated to the individual differences; specifically, we define the similarity as the overall maximum difference minus each individual difference: 
$\mbox{sim}_e = w_c (\delta_{max} - \delta_c(e)) + w_s (\delta_{max} - \delta_s(e) ) $, where $\delta_{max}$ is obtained by taking the max over all $\delta_c$ and $\delta_s$ in all edges from the similarity graph:  
$\delta_{max} = \max ( \max \delta_c (e), \max \delta_s (e))$. Therefore, 
the minimum similarity association rules can be derived by the following:
\begin{equation}
\begin{split}
    \min_{w_c, w_s} & \sum_{e \in G }
    \mbox{sim}_e^2 \\
     = & \sum_{e \in G} (w_c (\delta_{max} - \delta_c(e)) + w_s (\delta_{max} - \delta_s(e) ) )^2\\
    \textbf{s.t. } &  w_s + w_c = 1, w_s, w_c \geq 0 
\end{split}
\label{eq:optimization}
\end{equation}

Equation (\ref{eq:optimization}) is in the form of convex quadratic programming (QP) with only $2$ parameters. We use the off-the-shelf MATLAB {\it quadprog} solver to solve (\ref{eq:optimization}).
Not only is the solution efficient to obtain, but also this algorithm is easy to deploy. From retrospective data, we can calculate ${w}^\star_c, {w}^\star_s$  as the optimal solution from (\ref{eq:optimization}), and then the minimum similarity score can be calculated as the following: 
\begin{equation}
    \mbox{MSAR score: } r_{\mbox{\tiny{MSAR}}}^\star(v_i)  = {w}^\star_c \Tilde{c}(v_i) + {w}^\star_s \Tilde{s} (v_i). 
    \label{eq:score}
\end{equation}
The scores can be added directly to the learned rules along with confidence and support for each rule, and customers will be able to select top ranked rules based on MSAR scores. Table \ref{tab:msar_example} gives example of scores and how rules are compared during algorithm execution. 

\begin{table*}
\begin{subtable}{.80\linewidth}\centering
{
  \begin{tabular}{| c|| c|c| c | c |}
    \hline
        Comorbidities combinations candidates & Confidence & Support & AR score &	MSAR score (\ref{eq:score}) \\
        \hline
  1. Weight loss, Peripheral  & \multirow{ 2}{*}{0.681} &
   \multirow{ 2}{*}{0.00871} &    \multirow{ 2}{*}{{\textbf{0.681}}} & 
   \multirow{ 2}{*}{-0.192}\\
   vascular disease, Hypertension &&&& \\
   \hline
  2.  Weight loss, Peripheral  & \multirow{ 2}{*}{0.675} &
  \multirow{ 2}{*}{0.0244} &    \multirow{ 2}{*}{0.675} & \multirow{ 2}{*}{\textbf{0.0543}}\\
  vascular disease, Fluid-electrolyte disorders &&& & \\
  \hline
    \end{tabular}
}
\caption{Case I: the lower confidence rule has a support that is larger enough compared to a similar rule, and it is chosen by MSAR.}\label{tab:1a}
\end{subtable}%
\newline
\\
\\
\\
\begin{subtable}{.95\linewidth}\centering
{
  \begin{tabular}{| c|| c|c| c | c |}
    
    \hline
        Comorbidities combinations candidates & Confidence & Support & AR score &	MSAR score (\ref{eq:score})  \\
        \hline
 1. AIDS, Coagulopathy, Psychoses & 0.819 &
   0.000156 &    {\textbf{0.819}} & {\textbf{1.0014}} \\
  2.  AIDS, Coagulopathy, Renal failure &0.748 &
  0.000198 &    0.748 & 0.300\\
  \hline
    \end{tabular}

}

\caption{Case II: the lower confidence rule has only slightly greater support compared to a similar rule. AR and MSAR select the higher confidence rule.}\label{tab:1b}
\end{subtable}
\caption{MSAR balances confidence-support trade-offs with examples of comparison to the conventional association rules (AR) always selects the highest confidence rule, and therefore AR scores are the same as confidences. Bold texts indicate selected rules by AR and MSAR. }
\label{tab:msar_example}
\end{table*}

\section{Numeric results} 
We use the 
Banner Health EHR dataset to validate the proposed algorithm. It contains $8$ years of retrospective data from $30$ 
community teaching hospitals in the United States. 
 
\subsection{Statistics for all three recurrent patient criteria} 
At a population level, the combination of ED and inpatients produces a 30-day readmission rate of $12.3\%$,
 which is within a similar range to reported values on the HCUP dataset \cite{hcup}. Specifically, the rates of recurrent inpatient and ED visits 
are $7\%$ and $21\%$, respectively.

The percentage of total recurrent patients is about $25\%$ of all visits. 
Noticeably, ED recurrent patients contribute to $94\%$ of total recurrent patients (ED and inpatients combined). 

\subsection{MSAR results}
To obtain past comorbidities associated with recurrent/frequent visits, we first take ICD-9 and ICD-10 codes from a patient's past visits as training data. 
To avoid duplicate counts
, we only take the most recent (up to) $3$ visits for both recurrent and non-recurrent patients. One patient is therefore only counted once in the training data, and this process results in more than $400,000$ unique patients in 
that set.

The constant value in optimization equation (\ref{eq:optimization}) $\delta_{max}$ is obtained by taking the maximum of $z-$normalized confidences and supports, and the value is $4.34$ from our retrospective data. Optimal weight parameters of confidence and support are learned from retrospective data, and the numeric optimal solution 
is $w^\star_c = 0.778, w^\star_s = 0.221$. The weight of confidence is about $3.5$ times than the weight of support. We substitute these values into (\ref{eq:score}) to obtain MSAR rule scores for all candidate rules. 

\subsubsection{MSAR balances confidence-support trade-offs compared to conventional AR}
Table \ref{tab:msar_example} demonstrates MSAR and AR selections for similar rules while one has high-confidence low-support and the other one with low-confidence high-support.
In the top examples, the low confidence rule has larger support than the high confidence rule, and therefore MSAR picks the slighter lower confidence rule with much higher support. The bottom example illustrates the case when the low confidence rule has lower support than the high confidence rule, and MSAR agrees with the conventional highest confidence association rules. This behavior validated that \textbf{MSAR balances confidence-support trade-offs.}

\subsubsection{MSAR successfully selects challenging high-confidence, low-support comorbidities relates to recurrent visits}
\label{sub:pre}
\begin{figure}
    \centering
    \includegraphics[width = 8.5cm, height = 12.5cm]{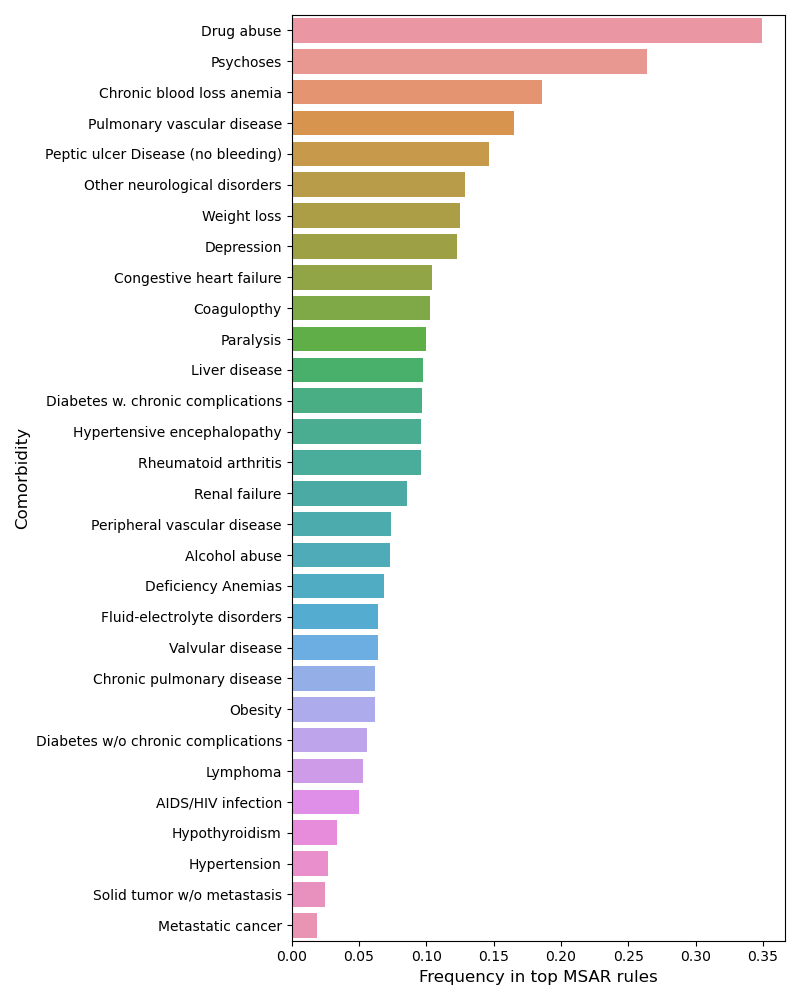}
    \caption{Frequency of comorbidities from the top $25\%$ of learned MSAR rules. Drug abuse, even with low prevalence compared to other comorbidities, has been successfully identified as being in $35\%$ of those top rules for recurrent visits.}
    \label{fig:top msar rules}
\end{figure}

In total $3860$ rules from triplets of Elixhauser comorbidities are learned. 
To understand the contributions of different comorbidities from learned rules, we take the top $1000$ $(25.9\%)$ rules and count the frequency of each comorbidity, which is depicted in Figure~\ref{fig:top msar rules}. 
Top categories include drug abuse, psychoses, neurological disorders, depression. Among them, drug abuse, reported as one of the top reasons for high utilization~\cite{stranges2011state}, even though with lower support, is identified by MSAR as a majority comorbidity associated with recurrent visits.

\begin{table}[]
    \centering
    \begin{tabular}{|c|c|}
\hline
Optimal weights & Mean (std) from CV
\\
\hline
$w_c^\star$ & 0.785 (0.0120)\\
$w_s^\star$ & 0.215 (0.0120)\\
\hline
\end{tabular}
 \caption{Distribution of optimal weights from $10-$fold 
 CV with $80\%$ randomly sampled training data in each fold. The standard deviations of optimal weights are fairly small, validating the consistency of learned weights by MSAR.}
 \label{tab:cv_weights}
\end{table}

\begin{table}[]
    \centering
    \begin{tabular}{|c|c|}
    \hline
        Algorithms & Mean (std) Rank-biased Overlap (RBO)~\cite{webber2010similarity} \\
        \hline
        MSAR & \textbf{0.973 (0.0138)} \\ 
        XGBoost\cite{chen2016xgboost} & 0.957 (0.0249) \\
        \hline
    \end{tabular}
    \caption{Pairwise RBOs for ranked comorbidities across $10$ folds. MSAR comorbidities are ranked by MSAR scores, and XGBoost comorbidities are ordered by Shapley values~\cite{shapley} on training data. MSAR has more consistent outputs.}
    \label{tab:rbo}
\end{table}

\begin{table}[]
    \centering
    \begin{tabular}{|c|c|}
    \hline
        Algorithms & $\%$ of indiscriminate across comorbidities\\
        \hline
        MSAR & \textbf{0$\%$} \\ 
        XGBoost\cite{chen2016xgboost} & $33.3\% - 40\%$ \\
        \hline
    \end{tabular}
    \caption{The percentages of comorbidities with zero weights. These can't be differentiated from each other while selecting top comorbidities associated with recurrent visits.}
    \label{tab:zero_ratios}
\end{table}

\begin{table*}[]
    \centering
    \begin{tabular}{|c|c| c|  c| c| }
    \hline
    \multirow{2}{*}{Comorbidity}
       & Frequency in  & Frequency in 
       & non-zero $\%$ of frequency & non-zero $\%$ of frequency  \\
        & top MSAR rules   & top-3 XGBoost features   & in top MSAR rules & in top-$3$ XGBoost features \\
        \hline
        Drug abuse & $0.35\  (0.003)$ &  $10^{-6}$ ($10^{-6}$) & \textbf{100$\%$} & 40$\%$ \\
        Weight loss& $0.13\ (0.004)$ & 5$\times 10^{-7}$ ($10^{-6}$)& \textbf{100}$\%$ & 10$\%$  \\
        AIDS & $0.049 \ (0.009)$
 & $0.0005 \ (0.0002)$ & \textbf{100$\%$}  &  \textbf{100}$\%$ \\
        \hline
    \end{tabular}
    \caption{Averages and standard deviations of frequency for example comorbidities, learned from MSAR rules, and non-zero percentage of frequency across 10 CV. When the frequency of a certain comorbidity is zero, it means that that comorbidity is not in any of top features associated with recurrent patients learned from algorithms. The non-zero percentage of frequency measures the percentage of frequency of certain comorbidity across 10 folds. MSAR is much more consistent in detecting low-support, high-confidence comorbidities than XGBoost.}
    \label{tab:zero_ratios_cv}
\end{table*}

\subsubsection{The consistency of learned weights and outputs of MSAR are validated through cross-validation (CV)}
To study the consistency of MSAR, we conduct a $10$-fold CV 
with $80\%$ of the training data in each fold. We first calculate optimal weights of confidence and support for each fold. Table~\ref{tab:cv_weights} shows the mean and standard deviation of optimal weights, 
which indicates low variance in optimal weights from various folds. 
Then, we rank the comorbidities by their MSAR scores in the same way as the previous section~\ref{sub:pre}. Finally, we calculate pair-wise Rank-based Overlap~\cite{webber2010similarity} from the outputs of $10$ folds, using ranked comorbidities lists ordered by MSAR scores. Table \ref{tab:rbo} shows that the pair-wised overlap across $10$ folds is very high with an average of $0.973$ overlap, validating the consistency of MSAR outputs.

\subsubsection{Compare with a popular XGBoost method, MSAR rules 
distinguishes across comorbidities better}
\label{subsec:xgboost_dis}
We compare MSAR with a decision tree-based, popular and commonly used method for tabular datasets, called XGBoost\cite{chen2016xgboost}. For the comparison, we use the same fold split as for MSAR. We use $30$ estimators, maximum depth at $6$ and $0.05$ as the learning rates, achieving 
on average $0.733$ area under the Receiver Operating Characteristic curve.

Although Table~\ref{tab:rbo} shows that the consistency of XGBoost across validations is lower than MSAR, the RBO overlap of outputs is fairly high. However, the main drawbacks of XGBoost for this task is the lack of interpretability and 
discrimination across different comorbidities. We investigated the frequency of comorbidities selected by the XGBoost model from its top $3$ rules and compared them with MSAR, whose numbers are reported in Table~\ref{tab:zero_ratios}. A large percentage of comorbidities, ranging from $33\%-40\%$ (corresponding to $10-12$ comorbidities) from $10-$fold CV, are not selected as top $3$ rules 
due to their low prevalences in training sets with around $330,000$ patients.  On the other hand, MSAR scores for all comorbidities are non-zero, making ranking possible for outputting top choices. \textbf{In summary, MSAR can distinguish all $30$ comorbidities; however $10-12$ comorbidities trained by XGBoost can't be discriminated.}

\subsubsection{Compare with XGBoost, MSAR rules select low-support high confidence comorbidities better}
\label{subsec:xgboost}
Table~\ref{tab:zero_ratios_cv} gives frequency from learned rules by MSAR and XGBoost of a few challenging comorbidities with low-support, high confidence. We can conclude that 
\textbf{MSAR is much more consistent at identifying challenging high-confidence low-support comorbidities compared to XGBoost}.  Additionally, 
{XGBoost missed low missed some high confidence however low support comorbidities such as drug abuse, weight loss.} Especially, {\textbf{drug abuse}} 
is reported in the literature as an important factor for recurrent patients,
{\textbf {even though having a high confidence, due to a lower support than some chronic conditions, XGBoost fails to select it as top $3$ contributor in $6$ out of $10$ folds, whereas MSAR successfully selects drug abuse out of all folds.}}

\section{Conclusion and Discussion}

In this work, we propose a framework for the healthcare system's need to identify recurrent patients.
 It identifies recurrent patients, is executable at admission, and finds the top comorbidities associated with their recurrent visits. This feat is accomplished via a novel proposed algorithm MSAR that balances the trade-off between confidence and support with optimal weights learnable from retrospective data. MSAR successfully learn high-confidence but low-support comorbidities, which is challenging for other algorithms such as XGBoost. The proposed framework can be used to assist future decisions to avoid future recurrent visits, such as recommending social-behavioral interventions and other relevant outpatient care. 
 
 The proposed MSAR algorithm has the 
 benefits of being customizable and easy to deploy in potential future products. Those who may find interest in such a proposal include leaders in the operations and population health space, along with front-line staff such as ED nurses and physicians, inpatient charge nurses, and patient flow coordinators.
This tool could help the care team
recommend potential interventions (e,g, rehabilitation, education programs, follow-up visits with primary care, medication monitoring, etc.) that help the patient reduce the risk of recurring ED and inpatient visits. 

\section*{Compliance with Ethical Standards}
Banner Health data use was a part of an ongoing retrospective deterioration detection study approved by the Institutional Review Board of Banner Health and by the Philips Internal Committee for Biomedical Experiments. Requirement for individual patient consent was waived because the project did not impact clinical care, was no greater than minimal risk, and all protected health information was deidentified to limited dataset.

\section*{Acknowledgment}
Thank Taiyao Wang for insightful discussions.\\
\bibliographystyle{IEEEtran} 
\bibliography{references_bibm}

\end{document}